\documentclass[conference]{IEEEtran}
\usepackage{caption}
\usepackage{subcaption}
\usepackage{nopageno}
\usepackage{graphbox} % it depends on "graphicx" package
\usepackage{amsmath,amsfonts,amsbsy,latexsym}
\usepackage{multirow}
\usepackage{makecell}
\usepackage{xcolor}
\usepackage{colortbl}
\usepackage{graphicx}
\usepackage{tabulary}
\usepackage{url}
\usepackage{booktabs}
\usepackage{cite}

\usepackage{smartdiagram}

\usepackage[activate={true,nocompatibility},
            final,
            tracking=true,
            kerning=true,
            spacing=true,
            factor=1100,
            stretch=10,
            shrink=10]{} %microtype

\usepackage[english]{babel}
\usepackage{blindtext}
\usepackage{helvet}
\usepackage[T1]{fontenc}
\usepackage{booktabs}
\usepackage{tabu}

\usepackage{amsmath}
\usepackage{amssymb}
\usepackage{amsthm}
\usepackage{algorithmic}
\usepackage{algorithm}
\usepackage{siunitx}
\usepackage{amsmath,amsfonts,amsthm,mathrsfs, mathtools}
\usepackage{multirow}
\usepackage{parallel}

\usepackage{marvosym}
\usepackage{graphicx}
\usepackage{pgfpages}
\graphicspath{{images/}{../images/}}
\usepackage{pgffor}
\usepackage{xcolor}
\usepackage{colortbl}
\usepackage{pgfplots}
\usepackage{tikz}
\usepackage{tikzsymbols}
\usetikzlibrary{automata,positioning, fadings,shadows, shapes.arrows, shapes.geometric}
\usetikzlibrary{matrix, chains, patterns}
\usetikzlibrary{arrows,shapes.gates.logic.US,shapes.gates.logic.IEC,calc,automata}
\usepgfplotslibrary{groupplots}

\usepackage{pifont}

\definecolor{Paired-2}{RGB}{166,206,227}
\definecolor{Paired-1}{RGB}{31,120,180}
\definecolor{Paired-4}{RGB}{178,223,138}
\definecolor{Paired-3}{RGB}{51,160,44}
\definecolor{Paired-6}{RGB}{251,154,153}
\definecolor{Paired-5}{RGB}{227,26,28}
\definecolor{Paired-8}{RGB}{253,191,111}
\definecolor{Paired-7}{RGB}{255,127,0}
\definecolor{Paired-10}{RGB}{202,178,214}
\definecolor{Paired-9}{RGB}{106,61,154}
\definecolor{Paired-12}{RGB}{255,255,153}
\definecolor{Paired-11}{RGB}{177,89,40}

\usepackage{tikz-timing}[2009/05/15]

\usepackage{scalerel} % needed package to scale the pdf-images perfectly

    \pgfdeclarelayer{background}
    \pgfdeclarelayer{foreground}
    \pgfsetlayers{background,main,foreground}
    \tikzstyle{vblock} = [draw,
                          fill=dateorange!30,
                          rotate=90,
                          anchor=north west,
                          minimum width=3.5cm,
                          minimum height=.5cm,
                         rounded corners=.05cm,
                          ]

    \tikzstyle{tile} = [draw,
                        fill=dateblue!30,
                        anchor=south west,
                        minimum width=3.5cm,
                        minimum height=.5cm,
                        rounded corners=.05cm,
                        ]

    \tikzstyle{nnlayer} = [draw,
                        fill=dateblue!30,
                        anchor=south west,
                        minimum width=1cm,
                        minimum height=.5cm,
                        rounded corners=.05cm,
                        ]

    \tikzstyle{petile} = [draw,
                        fill=dateorange!30,
                        anchor=south west,
                        minimum width=1.5cm,
                        minimum height=.5cm,
                        rounded corners=.05cm,
                        ]

    \tikzstyle{arrow} = [->,
                         >=stealth,
                         thick,
                         rounded corners=.1cm,
                         color=datemagenta,
                         ]

    \tikzstyle{arrowg} = [->,
                          >=stealth,
                          thick,
                          rounded corners=.1cm,
                          color=lightgray,
                          ]
    \tikzset{%
    table/.style={%
      matrix of nodes,
      row sep=-\pgflinewidth,
      column sep=-\pgflinewidth,
      nodes={rectangle,draw=black,text width=1.25ex,align=center},
      text depth=0.25ex,
      text height=1ex,
      nodes in empty cells
      },
    texto/.style={font=\footnotesize\sffamily},
    title/.style={font=\small\sffamily}
    }

    \pgfmathdeclarefunction{gauss}{2}{%
        \pgfmathparse{1/(#2*sqrt(2*pi))*exp(-((x-#1)^2)/(2*#2^2))}% chktex 36
    }

    \pgfmathdeclarefunction{gaussprune}{2}{%
        \pgfmathparse{or(x<-1,x>1)/(#2*sqrt(2*pi))*exp(-((x-#1)^2)/(2*#2^2))}% chktex 36
    }

\usepackage{stfloats}

\usepackage{perpage}
\usepackage{hyperref}
\usepackage{csquotes}

\usepackage{pgfpages}
\makeatother

\definecolor{dateblue}{RGB}{36,80,117}
\definecolor{datemagenta}{RGB}{183,50,101}
\definecolor{dateorange}{RGB}{198,84,54}
\definecolor{datebrown}{RGB}{198,140,54}
\definecolor{dateyellow}{RGB}{198,178,54}

\tikzfading[name=arrowfading, top color=transparent!0, bottom color=transparent!95]
\tikzset{arrowfill/.style={#1, general shadow={fill=black, shadow yshift=-0.8ex, path fading=arrowfading}}}
\tikzset{arrowstyle/.style n args={3}{draw=#2,arrowfill={#3}, single arrow,minimum height=#1, single arrow,
single arrow head extend=.3cm,}}

\NewDocumentCommand{\tikzfancyarrow}{O{2cm} O{dateblue} O{top color=dateblue!20, bottom color=dateblue} m}{
\tikz[baseline=-0.5ex]\node [arrowstyle={#1}{#2}{#3}] {#4};
}

\usepackage{circuitikz}

{
\begin{tikzpicture}[baseline=(current bounding box.north),scale=0.8]
\draw (0,0) grid (4,4);
\draw (0,4) -- node [pos=0.7,above right,anchor=south west] {$x_3x_4$} node [pos=0.7,below left,anchor=north east] {$x_1x_2$} ++(135:1);
\matrix (mapa) [matrix of nodes,
        column sep={0.8cm,between origins},
        row sep={0.8cm,between origins},
        every node/.style={minimum size=0.3mm},
        anchor=8.center,
        ampersand replacement=\&] at (0.5,0.5)
{
                       \& |(c00)| 00         \& |(c01)| 01         \& |(c11)| 11         \& |(c10)| 10         \& |(cf)| \phantom{00} \\
|(r00)| 00             \& |(0)|  \phantom{0} \& |(1)|  \phantom{0} \& |(3)|  \phantom{0} \& |(2)|  \phantom{0} \&                     \\
|(r01)| 01             \& |(4)|  \phantom{0} \& |(5)|  \phantom{0} \& |(7)|  \phantom{0} \& |(6)|  \phantom{0} \&                     \\
|(r11)| 11             \& |(12)| \phantom{0} \& |(13)| \phantom{0} \& |(15)| \phantom{0} \& |(14)| \phantom{0} \&                     \\
|(r10)| 10             \& |(8)|  \phantom{0} \& |(9)|  \phantom{0} \& |(11)| \phantom{0} \& |(10)| \phantom{0} \&                     \\
|(rf) | \phantom{00}   \&                    \&                    \&                    \&                    \&                     \\
};
}%
{
\end{tikzpicture}
}

{
\begin{tikzpicture}[baseline=(current bounding box.north),scale=0.8]
\draw (0,0) grid (4,2);
\draw (0,2) -- node [pos=0.7,above right,anchor=south west] {$x_2x_3$} node [pos=0.7,below left,anchor=north east] {$x_1$} ++(135:1);
\matrix (mapa) [matrix of nodes,
        column sep={0.8cm,between origins},
        row sep={0.8cm,between origins},
        every node/.style={minimum size=0.3mm},
        anchor=4.center,
        ampersand replacement=\&] at (0.5,0.5)
{
                      \& |(c00)| 00         \& |(c01)| 01         \& |(c11)| 11         \& |(c10)| 10         \& |(cf)| \phantom{00} \\
|(r00)| 0             \& |(0)|  \phantom{0} \& |(1)|  \phantom{0} \& |(3)|  \phantom{0} \& |(2)|  \phantom{0} \&                     \\
|(r01)| 1             \& |(4)|  \phantom{0} \& |(5)|  \phantom{0} \& |(7)|  \phantom{0} \& |(6)|  \phantom{0} \&                     \\
|(rf) | \phantom{00}  \&                    \&                    \&                    \&                    \&                     \\
};
}%
{
\end{tikzpicture}
}

{
\begin{tikzpicture}[baseline=(current bounding box.north),scale=0.8]
\draw (0,0) grid (4,2);
\draw (0,2) -- node [pos=0.7,above right,anchor=south west] {$x_1x_2$} node [pos=0.7,below left,anchor=north east] {$x_5$} ++(135:1);
\matrix (mapa) [matrix of nodes,
        column sep={0.8cm,between origins},
        row sep={0.8cm,between origins},
        every node/.style={minimum size=0.3mm},
        anchor=4.center,
        ampersand replacement=\&] at (0.5,0.5)
{
                      \& |(c00)| 00         \& |(c01)| 01         \& |(c11)| 11         \& |(c10)| 10         \& |(cf)| \phantom{00} \\
|(r00)| 0             \& |(0)|  \phantom{0} \& |(1)|  \phantom{0} \& |(3)|  \phantom{0} \& |(2)|  \phantom{0} \&                     \\
|(r01)| 1             \& |(4)|  \phantom{0} \& |(5)|  \phantom{0} \& |(7)|  \phantom{0} \& |(6)|  \phantom{0} \&                     \\
|(rf) | \phantom{00}  \&                    \&                    \&                    \&                    \&                     \\
};
}%
{
\end{tikzpicture}
}

{
\begin{tikzpicture}[baseline=(current bounding box.north),scale=0.8]
\draw (0,0) grid (4,1);
\draw (0,1) -- node [pos=0.7,above right,anchor=south west] {$x_1x_2$} node [pos=0.7,below left,anchor=north east] {} ++(135:1);
\matrix (mapa) [matrix of nodes,
        column sep={0.8cm,between origins},
        row sep={0.7cm,between origins},
        every node/.style={minimum size=0.3mm},
        anchor=5.center,
        ampersand replacement=\&] at (0.5,0.8)
{
 \& |(c00)| 00         \& |(c01)| 01         \& |(c10)| 10         \& |(c11)| 11         \& |(cf)| \phantom{00} \\
 |(rf) | \phantom{00}   \& |(0)|  \phantom{0} \& |(1)|  \phantom{0} \& |(2)|  \phantom{0} \& |(3)|  \phantom{0} \&                     \\
};
}%
{
\end{tikzpicture}
}
{
\begin{tikzpicture}[baseline=(current bounding box.north),scale=0.8]
\draw (0,0) grid (4,1);
\draw (0,1) -- node [pos=0.7,above right,anchor=south west] {$x_1x_2$} node [pos=0.7,below left,anchor=north east] {} ++(135:1);
\matrix (mapa) [matrix of nodes,
        column sep={0.8cm,between origins},
        row sep={0.7cm,between origins},
        every node/.style={minimum size=0.3mm},
        anchor=5.center,
        ampersand replacement=\&] at (0.5,0.8)
{
 \& |(c00)| 00         \& |(c01)| 01         \& |(c11)| 11         \& |(c10)| 10         \& |(cf)| \phantom{00} \\
 |(rf) | \phantom{00}   \& |(0)|  \phantom{0} \& |(1)|  \phantom{0} \& |(3)|  \phantom{0} \& |(2)|  \phantom{0} \&                     \\
};
}%
{
\end{tikzpicture}
}

{
\begin{tikzpicture}[baseline=(current bounding box.north),scale=0.8]
\draw (0,0) grid (2,2);
\draw (0,2) -- node [pos=0.7,above right,anchor=south west] {$x_2$} node [pos=0.7,below left,anchor=north east] {$x_1$} ++(135:1);
\matrix (mapa) [matrix of nodes,
        column sep={0.8cm,between origins},
        row sep={0.8cm,between origins},
        every node/.style={minimum size=0.3mm},
        anchor=2.center,
        ampersand replacement=\&] at (0.5,0.5)
{
          \& |(c00)| 0          \& |(c01)| 1  \\
|(r00)| 0 \& |(0)|  \phantom{0} \& |(1)|  \phantom{0} \\
|(r01)| 1 \& |(2)|  \phantom{0} \& |(3)|  \phantom{0} \\
};
}%
{
\end{tikzpicture}
}

\usepackage{listings}
\usetikzlibrary{positioning, shapes, arrows.meta}

\usepackage[utf8x]{inputenc}

\definecolor{shellGreen}{RGB}{19,193,106}
\definecolor{backcolor}{rgb}{0.95,0.95,0.92}
\definecolor{mateBlack}{RGB}{45,45,50}
\definecolor{comment}{rgb}{0.1,0.6,0.2}
\definecolor{codegray}{rgb}{0.5,0.5,0.5}

\lstdefinestyle{verilog}{
   language=verilog,
   frame=single,
   basicstyle=\ttfamily\footnotesize,
   breaklines=true,
   captionpos=t,
   keepspaces=true,
   backgroundcolor=\color{backcolor},
   keywordstyle=[1]\color{blue}\bf,
   keywordstyle=[2]\color{red}\bf,
   keywordstyle=[3]\color{cyan!50}\bf,
   stringstyle=\color{orange},
   commentstyle=\color{comment},
   tabsize=2,
   showspaces=false,
   showstringspaces=false,
   showtabs=false,
   morekeywords=[1]{
      library, use ,all,entity,is,port,in,out,end,architecture,of, body,
      function, variable, begin,and,or,Not,downto,ALL, signal, process, if,
      else, elsif, case, when, then, range, to, component, type, with, select,
      others, constant, inout, buffer, map, true, false, array, subtype, wait,
      wait for, generic, =, <, >, <=, >=, =>, 
   },
   alsoletter={=, <, >},
   morekeywords=[2]{
          STD, textio, std_logic_textio, STD_LOGIC_VECTOR,STD_LOGIC,IEEE,STD_LOGIC_1164, work, local, real,
          math_real, time, NUMERIC_STD,STD_LOGIC_ARITH,STD_LOGIC_UNSIGNED,
          std_logic_vector, std_logic, ieee, numeric_std, std_ulogic,
          std_logic_1164, natural, bit, bit_vector, signed, unsigned,
          boolean, integer
    },
    morekeywords=[3]{rising_edge, falling_edge, resize, to_signed, to_unsigned},
    morecomment=[l]{--},
    rulecolor=\color{black},
}

\usepackage{diagbox}

\definecolor{codegreen}{rgb}{0,0.6,0}
\definecolor{codegray}{rgb}{0.5,0.5,0.5}
\definecolor{codepurple}{rgb}{0.58,0,0.82}
\definecolor{backcolour}{rgb}{0.95,0.95,0.92}

\lstdefinestyle{mystyle}{
    backgroundcolor=\color{backcolour}\bf,   
    commentstyle=\color{codegreen}\bf,
    keywordstyle=\color{magenta}\bf,
    numberstyle=\tiny\color{codegray}\bf,
    stringstyle=\color{codepurple},
    basicstyle=\ttfamily\footnotesize,
    breakatwhitespace=false,         
    breaklines=true,                 
    captionpos=t,                    
    keepspaces=true,                
    showspaces=false,                
    showstringspaces=false,
    showtabs=false,                  
    tabsize=2
}

\definecolor{vscode-bg}{HTML}{1E1E1E}      % Background color
\definecolor{vscode-text}{HTML}{D4D4D4}    % Default text color
\definecolor{vscode-keyword}{HTML}{569CD6} % Keywords (blue)
\definecolor{vscode-string}{HTML}{D69D85}  % Strings (orange)
\definecolor{vscode-comment}{HTML}{6A9955} % Comments (green)
\definecolor{vscode-function}{HTML}{DCDCAA}% Function name (yellow)
\definecolor{vscode-number}{HTML}{B5CEA8}  % Numbers (light green)

\lstdefinestyle{vscode}{
    backgroundcolor=\color{vscode-bg},
    basicstyle=\ttfamily\footnotesize\color{vscode-text},
    keywordstyle=\color{vscode-keyword}\bfseries,
    commentstyle=\color{vscode-comment}\itshape,
    stringstyle=\color{vscode-string},
    numberstyle=\color{vscode-number},
    identifierstyle=\color{vscode-text},
    morekeywords={self}, % Add Python keywords or any language-specific ones
    numbers=left,
    numberstyle=\tiny\color{gray},
    numbersep=5pt,
    frame=single,
    rulecolor=\color{gray},
    breaklines=true,
    showstringspaces=false,
    tabsize=4,
    captionpos=b,
    keepspaces=true,
    columns=fullflexible,
}

\definecolor{vscode-bg}{HTML}{FFFFFF}      % Light background color
\definecolor{vscode-text}{HTML}{000000}    % Default text color (black)
\definecolor{vscode-keyword}{HTML}{0000FF} % Keywords (blue)
\definecolor{vscode-string}{HTML}{A31515}  % Strings (red)
\definecolor{vscode-comment}{HTML}{008000} % Comments (green)
\definecolor{vscode-function}{HTML}{795E26}% Function name (brownish)
\definecolor{vscode-number}{HTML}{098658}  % Numbers (greenish)

\lstdefinestyle{vscode-light}{
    backgroundcolor=\color{vscode-bg},
    basicstyle=\ttfamily\footnotesize\color{vscode-text},
    keywordstyle=\color{vscode-keyword}\bfseries,
    commentstyle=\color{vscode-comment}\itshape,
    stringstyle=\color{vscode-string},
    numberstyle=\color{vscode-number},
    identifierstyle=\color{vscode-text},
    morekeywords={self}, % Add Python-specific keywords
    numbers=left,
    numberstyle=\tiny\color{gray},
    numbersep=5pt,
    frame=single,
    rulecolor=\color{gray},
    breaklines=true,
    showstringspaces=false,
    tabsize=4,
    captionpos=b,
    keepspaces=true,
    columns=fullflexible,
    mathescape=true % Enable math mode using $...$
}

\definecolor{cnf-keyword}{RGB}{0,0,255}   % Blue for 'p cnf'
\definecolor{cnf-comment}{RGB}{0,128,0}   % Green for comments
\definecolor{cnf-number}{RGB}{255,0,0}    % Red for literals (numbers)
\definecolor{cnf-operator}{RGB}{0,0,0}    % Black for the '0' and negative sign

\lstdefinestyle{cnf}{
    backgroundcolor=\color{white},
    basicstyle=\ttfamily\footnotesize,
    keywordstyle=\color{cnf-keyword}\bfseries,
    commentstyle=\color{cnf-comment},
    numberstyle=\color{gray},
    numbersep=5pt,
    frame=single,
    breaklines=true,
    showstringspaces=false,
    tabsize=4,
    columns=fullflexible,
    morekeywords={p, cnf},          % 'p' and 'cnf' are treated as keywords
    morecomment=[l]{c},             % Comments start with 'c'
    morestring=[b][\color{cnf-number}]0,  % Highlight literals ending with '0'
    alsoletter={-},                 % Treat '-' as part of numbers
    literate=%
    {-}{{\textcolor{cnf-operator}{-}}}1, % Highlight negative signs
    mathescape=true
}
\usepackage[a4paper, total={184mm,262mm}]{geometry} %184 239
\def\BibTeX{{\rm B\kern-.05em{\sc i\kern-.025em b}\kern-.08em
    T\kern-.1667em\lower.7ex\hbox{E}\kern-.125emX}}
\begin{document}
\bstctlcite{IEEEexample:BSTcontrol}
% \title{\Huge{Inherently-Parallel Differential Sampling of Multi-Level Digital Circuits}}

\title{High-Throughput SAT Sampling\vspace{-0.25cm}}

% \title{Conference Paper Title*\\
% {\footnotesize \textsuperscript{*}Note: Sub-titles are not captured in Xplore and
% should not be used}
% \thanks{Identify applicable funding agency here. If none, delete this.}
% }
%%%%%%%%%%%%%%%%%%%%%%%%%%%%%%%%%%%%%%%%%%%%%%%%%%%%%%%%
% \author{\IEEEauthorblockN{
%     Arash Ardakani, 
%     Minwoo Kang,
%     Kevin He,
%     Qijing Huang,
%     John Wawrzynek
% }
% }

\makeatletter
    \newcommand{\linebreakand}{%
      \end{@IEEEauthorhalign}
      \hfill\mbox{}\par
      \mbox{}\hfill\begin{@IEEEauthorhalign}
    }
    \makeatother

\author{
    \IEEEauthorblockN{Arash Ardakani}
    \IEEEauthorblockA{
    \textit{University of California, Berkeley}\\
    arash.ardakani@berkeley.edu}
    \and
    \IEEEauthorblockN{Minwoo Kang}
    \IEEEauthorblockA{
    \textit{University of California, Berkeley}\\
    minwoo\_kang@berkeley.edu}
    \and
    \IEEEauthorblockN{Kevin He}
    \IEEEauthorblockA{
    \textit{University of California, Berkeley}\\
    kevinjhe@berkeley.edu}
    \linebreakand
    \IEEEauthorblockN{Qijing Huang}
    \IEEEauthorblockA{
    \textit{NVIDIA}\\
    jennyhuang@nvidia.com}
    \and
    \IEEEauthorblockN{John Wawrzynek}
    \IEEEauthorblockA{
    \textit{University of California, Berkeley}\\
    johnw@berkeley.edu}
    }
    
\makeatletter
\patchcmd{\@maketitle}
  {\addvspace{0.5\baselineskip}\egroup}
  {\addvspace{-2\baselineskip}\egroup}
  {}
  {}
\makeatother

%%%%5%%%%%%%%%%%%%%%%%%%%%%%%%%%%%%%%%%%%%%%%%%%%%%%%%%%%

\newcommand\blfootnote[1]{%
  \begingroup
  \renewcommand\thefootnote{}\footnote{#1}%
  \addtocounter{footnote}{-1}%
  \endgroup
}

\definecolor{main}{HTML}{4472C4}    % setting main color to be used
\definecolor{sub}{HTML}{EBF4FF}     % setting sub color to be used
\newcommand{\com}[1]{{\color{red}\sf{[#1]}}}
\newcommand{\OURS}{{\sc Demotic}}
%%%%%%%%%%%%%%%%%%%%%%%%%%%%%%%%%%%%%%%%%%%%%%%%%%%%%%%%

% \input{_0_abstract}

\maketitle

\begin{abstract}
  In this work, we present a novel technique for GPU-accelerated Boolean satisfiability (SAT) sampling. Unlike conventional sampling algorithms that directly operate on conjunctive normal form (CNF), our method transforms the logical constraints of SAT problems by factoring their CNF representations into simplified multi-level, multi-output Boolean functions. It then leverages gradient-based optimization to guide the search for a diverse set of valid solutions. Our method operates directly on the circuit structure of refactored SAT instances, reinterpreting the SAT problem as a supervised multi-output regression task. This differentiable technique enables independent bit-wise operations on each tensor element, allowing parallel execution of learning processes. As a result, we achieve GPU-accelerated sampling with significant runtime improvements ranging from $33.6\times$ to $523.6\times$ over state-of-the-art heuristic samplers. We demonstrate the superior performance of our sampling method through an extensive evaluation on $60$ instances from a public domain benchmark suite utilized in previous studies.

\end{abstract}

\vspace{-0.25cm}
\section{Introduction}

High-throughput SAT samplers play a crucial role in advancing the state of the art in software and hardware verification methodologies \cite{dutra2018quicksampler}.
Generating a set of random solutions to logical constraints is critical in the verification, testing, and synthesis.
In software verification, SAT samplers enable efficient exploration of diverse execution paths, addressing the scalability challenges inherent in symbolic execution~ \cite{Clarke1976testing, King1976symbolic, Avgerinos2014dynamic, Anand2011test, Anand2007test, Artzi2008test, Burnim2008test, Cadar2008test, Chipounov2012test, Godefroid2005test, Jayaraman2009jFuzzAC, yoshida2017test, corina2008test, saxena2010test, dawn2008test, Tillmann2008test}. In hardware verification, they support the generation of varied input patterns, ensuring a rigorous and effective verification process~\cite{Kitchen2007crv, Zhao2009crv, Naveh2013crv, Naveh2006crv}.

The SAT sampling process begins by formulating the logical constraints of the target application into conjunctive normal form (CNF) \cite{Biere2009SAT}.
CNF is the specific format required by most SAT samplers, where the logical formula is expressed as a conjunction of clauses, with each clause being a disjunction of literals.
%Therefore, converting logical relationships into CNF is a critical step for effectively using SAT samplers. 
%This conversion ensures that the resulting CNF accurately represents the original behavior of the target application.
The complexity of the logical constraints in the target application can result in a CNF that is not always concise. Nevertheless, CNF remains the preferred format due to the strong performance of SAT samplers and solvers. The complexity of the CNF can, however, affect the efficiency of these solvers.

SAT solvers employ various strategies to find a satisfying assignment for the variables in the CNF. Modern SAT solvers \cite{Niklas2003SAT, Moskewicz2001Chaff, Audemard2018Glucose} often use the conflict-driven clause learning (CDCL) algorithm \cite{Silva1996CDCL, silva2021CDCL}, that relies heavily on heuristics such as conflict-driven backtracking and clause learning. These heuristics effectively guide the CDCL algorithm in finding a satisfying assignment. Due to the sequential nature of these heuristics, that involve branching and backtracking, the latest SAT solvers are typically executed on CPUs. Consequently, state-of-the-art SAT samplers, which incorporate SAT solvers within their algorithms, also rely on a sequential process and are optimized for CPU execution.

%GPU acceleration has shown significant performance improvements over CPUs in a wide range of applications, particularly in machine learning \cite{Krizhevsky2012AlexNet}.
%GPU computing is well-suited to problems that involve large amounts of data parallelism.
Generating multiple satisfying solutions to the SAT problem is a good match to GPU computing, provided that a sampling method is available that performs consistent, data-parallel computations.
To address this opportunity, we propose a transformation algorithm that converts logical constraints encoded as a CNF into a simplified multi-level, multi-output Boolean function while maintaining the original logical constraints. This transformation significantly reduces the number of bit-wise operations and thus lowers the complexity of the sampling process.
We then formulate the resulting simplified multi-level, multi-output Boolean function as a multi-output regression task, where each logic gate is represented probabilistically, enabling the use of gradient descent (GD) to learn diverse solutions. This approach enables the parallel generation of independent SAT problem solutions, allowing for effective GPU acceleration.
We demonstrate the superior performance of our sampling method across $60$ instances from a publicly available benchmark suite \cite{meel_2020_benchmark} used in previous studies. The code of our sampler is available at \url{https://github.com/arashardakani/High-Throughput-SAT-Sampler}.

\vspace{-0.2cm}
\section{Preliminaries}
\subsection{SAT Sampling} \label{sec:sat-sampling}

SAT sampling involves drawing solutions from the solution space defined by a set of logical constraints expressed in CNF. In SAT sampling applications, Boolean expressions are typically represented in higher-level logical formats before being converted into CNF \cite{Biere2009SAT, Barrett2013SAT}. These formats include propositional logic with operators like AND, OR, NOT, implications, and equivalences, as well as more complex structures such as if-then-else conditions, arithmetic expressions, and bit-level operations. In hardware verification, Boolean expressions can take the form of circuit representations, such as And-Inverter Graphs (AIGs) or Binary Decision Diagrams (BDDs). In cryptographic contexts, Algebraic Normal Form (ANF) is sometimes used. These representations are transformed into CNF through logical simplifications, flattening complex structures, and applying techniques like Tseitin transformation \cite{tseitin1983complexity}. This transformation preserves the satisfiability of the original formula while introducing auxiliary variables when needed. The conversion to CNF provides SAT solvers with a standardized problem representation that retains the essential constraints of the original problem.

% As discussed in Section \ref{sec:sat-sampling}, Boolean expressions are typically expressed in more abstract logical formats before being converted to CNF through techniques like logical simplifications and transformations such as the Tseitin transformation. This transformation preserves the satisfiability of the original formula while introducing auxiliary variables when needed. The conversion to CNF provides SAT solvers with a standardized problem representation that retains the essential constraints of the original problem.

A CNF consists of a conjunction of clauses (i.e., an AND of multiple clauses), where each clause consists of a disjunction of literals (i.e., an OR of literals). Literals refer to Boolean variables or their negations. In SAT solving, the goal is to determine if there exists an assignment of binary values to the variables in a given CNF, representing a Boolean expression, such that all clauses evaluate to $1$. SAT sampling adds a probabilistic layer to this process. Instead of seeking just one solution for satisfiable instances, the aim is to produce multiple solutions or samples from the complete set of possible solutions. Generating samples from SAT instances plays a crucial role in design verification, testing, and synthesis, with significant applications in constrained-random verification (CRV) \cite{Kitchen2007crv}.

A common method for SAT sampling involves using SAT solvers with built-in sampling functionality. These solvers are designed not only to determine the satisfiability of a Boolean formula but also to extract solutions from the solution space. Efficient SAT solving techniques include backtracking algorithms like the Davis-Putnam-Logemann-Loveland (DPLL) algorithm \cite{Davis1962DPLL}, stochastic local search methods such as WalkSAT \cite{selman1993local}, and CDCL algorithms \cite{Silva1996CDCL, silva2021CDCL}. In recent years, several approaches have been developed for SAT sampling, including randomized algorithms, Markov chain Monte Carlo (MCMC) techniques, and heuristic-based sampling methods \cite{Impagliazzo2017RandomSAT, kitchen2009markov, Soos2020unigen3, dutra2018quicksampler, Golia2021cmsgen}. These methods typically explore the solution space iteratively, selecting candidate solutions based on predefined criteria and stochastically deciding whether to accept or reject them.

SAT solvers and samplers have been optimized over decades to efficiently handle problems in CNF. CNF is well-suited for SAT-solving algorithms like DPLL and CDCL. These algorithms take advantage of CNF's structure to systematically explore possible truth assignments, detect conflicts early, and prune the search space efficiently. By focusing on individual clauses, which define specific constraints on solutions, the use of CNF enables solvers to address highly complex problems in a manageable way.

\subsection{Multi-Output Regression Task}
A multi-output regression task, in the context of example generation, refers to the process of creating data points (or ``examples'') that serve as input-output pairs for a given model \cite{Ardakani2024diffsampler}. The main goal is to generate inputs such that they satisfy multiple specific output constraints simultaneously. Various methods, such as linear regression, neural networks or probabilistic representations of the relationships between inputs and outputs, can be used to construct such a model. The inputs to the model are adjusted in an iterative manner to minimize the difference between the predicted and actual output values. One common way to measure this difference is by using metrics like mean squared error (MSE) or $\ell_2$ loss.

\section{Methodology}
% SAT solvers and samplers have been optimized over decades to efficiently handle problems in CNF. CNF, a form that consists of a conjunction of disjunctions (i.e., clauses), is well-suited for SAT-solving algorithms like DPLL and CDCL. These algorithms take advantage of CNF's structure to systematically explore possible truth assignments, detect conflicts early, and prune the search space efficiently. By focusing on individual clauses, which define specific constraints on solutions, the use of CNF enables solvers to address highly complex problems in a manageable way.

% As discussed in Section \ref{sec:sat-sampling}, Boolean expressions are typically expressed in more abstract logical formats before being converted to CNF through techniques like logical simplifications and transformations such as the Tseitin transformation. This transformation preserves the satisfiability of the original formula while introducing auxiliary variables when needed. The conversion to CNF provides SAT solvers with a standardized problem representation that retains the essential constraints of the original problem.

In this section, we present a transformation algorithm paired with an optimization method to generate valid and diverse solutions to SAT problems. Our algorithm transforms a CNF from its flat, two-level structure into a more streamlined, multi-level, multi-output Boolean function, reducing both operations and logical constraints. We then apply gradient-based optimization to efficiently learn valid solutions using the simplified multi-level, multi-output function.

As discussed in Section \ref{sec:sat-sampling}, Boolean expressions are typically represented in more abstract logical formats before being converted to CNF. In other words, the sub-expressions (i.e., sub-clauses) in a CNF often result from the transformation of single logical operators or the combination of multiple operators. This transformation increases the size of the CNF. Therefore, a sampler that could operate directly on the logical operators represented by the sub-expressions in a CNF would be advantageous. Motivated by this, we introduce our transformation algorithm, which converts CNF into an equisatisfiable multi-level, multi-output Boolean function, and demonstrate how this function can be used to generate various valid and distinct solutions using gradient-based optimization.

\subsection{Transformation Algorithm}
Let us first review the clauses, known as the CNF signature \cite{roy2004restoring}, which represent primary logical operators as a result of the Tseitin transformation. The CNF structure of an inverter with the input $x$ and the output $f$, i.e., $f(x) = \neg x$, is given by 
\begin{equation}
    (f \vee x) \wedge (\neg f \vee \neg x), 
\end{equation} 
where $\vee$ and $\wedge$ denote logical OR and AND operators, respectively. The clauses representing the logical OR operator with $n$ inputs and an output $f$, i.e., $f(x_1, x_2, \dots, x_n) = \bigvee_{i=1}^n x_i$, can be expressed as
\begin{equation}
    \left(\neg f \vee \bigvee_{i=1}^n x_i \right) \wedge \left( \bigwedge_{i=1}^n (f \vee \neg x_i) \right).
\end{equation}
The clauses representing the logical AND operator with $n$ inputs and an output $f$, i.e., $f(x_1, x_2, \dots, x_n) = \bigwedge_{i=1}^n x_i$, are similar to those of the logical OR operator with its input and output variables inverted, i.e., 
\begin{equation}
    \left(f \vee \bigvee_{i=1}^n \neg x_i \right) \wedge \left( \bigwedge_{i=1}^n (\neg f \vee x_i) \right).
\end{equation}
The CNF structure of the logical NAND and NOR operators can be derived in a similar manner by inverting the output variable in the clauses associated with the logical AND and OR operators, respectively. The CNF signature of the logical XOR operator with $n$ inputs and an output $f$ is is given by \vspace{-0.2cm}
\begin{align}
    & (\neg f \vee \text{XOR}(x_1, \dots, x_n)) \wedge (f \vee \neg \text{XOR}(x_1, \dots, x_n)) \nonumber \\
    = & \neg \text{XOR}(x_1, \dots, x_n, f) = \text{XNOR}(x_1, \dots, x_n, f).\vspace{-0.2cm}
\end{align}
Similarly, the CNF structure for the logical XNOR operation can be described as $\text{XOR}(x_1, \dots, x_n, f)$.

While deriving logical operators from the CNF signatures described above is straightforward through pattern matching, this method is limited to identifying clauses linked to primary logical operators. It does not handle sub-expressions that may involve more complex Boolean expressions constructed from these operators. For example, the clauses\vspace{-0.2cm}
\begin{align}
    & (\neg x_4 \vee \neg x_{107} \vee x_5) \wedge (\neg x_4 \vee x_{107} \vee \neg x_5) \wedge \nonumber \\
    & (x_4 \vee \neg x_{108} \vee x_5) \wedge (x_4 \vee x_{108} \vee \neg x_5) \vspace{-0.2cm}
    \label{c1}
\end{align}
from the '$75$-$10$-$1$-q' CNF instance correspond to the function $x_5(x_4, x_{107}, x_{108}) = (x_{107} \wedge x_4) \vee (x_{108} \wedge \neg x_4)$, which cannot be identified using pattern matching alone, as it is impractical to store all possible Boolean patterns. This underscores the need for a general transformation algorithm capable of identifying the Boolean sub-expressions and constraints represented by the clauses.

Before introducing our algorithm, let us establish some key definitions. The primary objective of the algorithm is to transform a given CNF into an acyclic combinational structure while preserving all logical constraints in the original CNF. Variables that correspond to the inputs and outputs of this circuit are referred to as primary input variables and primary output variables, respectively. Variables representing the intermediate signals within the circuit are called intermediate variables. With these definitions in place, once a variable is identified as a primary input or intermediate variable, it cannot be redefined as an output variable in subsequent Boolean expressions, due to the acyclic nature of the circuit.

The main challenge in this transformation is first to identify sub-clauses corresponding to a Boolean expression and then to derive that expression. To achieve this, we begin by reading the first clause in the CNF. For each variable in the clause, if it has not already been defined as a primary input or intermediate variable, we treat it as a potential output of the clause or clauses read so far. We then derive its Boolean expression based on the clause or clauses (read so far) containing this variable in its negated form, and similarly derive its complementary expression from the clauses where the variable appears in its true form. If the resulting Boolean expressions are indeed complements of each other, we have successfully identified the Boolean expression for the current set of clauses. If not, we proceed to the next clause and repeat the process until the corresponding Boolean expression is found. Once the Boolean expression is identified, we designate its output variable as an intermediate variable. Additionally, if any input variables in the derived Boolean expression have not yet been classified as intermediate variables, they are now recognized as primary input variables. In case the resulting Boolean expression is identified to be a constant Boolean function with its output being a constant value of either $0$ or $1$, its output variable is recognized as a primary output variable. The obtained Boolean expression is simplified before adoption in the final circuit structure. For all Boolean manipulations, such as simplification and complement checking, we utilize the symbolic Boolean algebra system SymPy \cite{sympy}, a Python library for symbolic mathematics. It is worth noting that the resulting multi-level, multi-output Boolean function from our transformation can be further optimized by leveraging other techniques \cite{Robert2010ABC, Alan2006synthesis, Alan2011synthesis}, for reducing the complexity of multi-level logic circuits, potentially leading to even more compact Boolean functions. Our transformation process is summarized in Algorithm \ref{alg1}.

To clarify the transformation process, let us revisit the clauses in Eq. (\ref{c1}). Treating $x_5$ as a potential output variable, we derive its Boolean expression from the clauses where $x_5$ appears in its negated form (i.e., $(\neg x_4 \vee x_{107} \vee \neg x_5) \wedge (x_4 \vee x_{108} \vee \neg x_5)$). Since $\neg x_5 = 0$ in these clauses, it becomes non-contributory, resulting in the Boolean expression $x_5(x_4, x_{107}, x_{108}) = (x_{107} \wedge x_4) \vee (x_{108} \wedge \neg x_4)$. This is because we are determining the Boolean expression for $x_5 = 1$, and thus the remaining clauses containing $x_5$ in its true form are already satisfied and do not contribute to the expression.

Similarly, we derive the complementary Boolean expression using the clauses where $x_5$ appears in its true form (i.e., $(\neg x_4 \vee \neg x_{107} \vee x_5) \wedge (x_4 \vee \neg x_{108} \vee x_5)$). In this case, $x_5 = 0$ rendering it non-contributory and resulting in $\neg x_5(x_4, x_{107}, x_{108}) = (\neg x_{107} \wedge x_4) \vee (\neg x_{108} \wedge \neg x_4)$. Since these two expressions are complementary, it confirms the validity of the Boolean expression with the specified input and output variables. It is worth mentioning that repeating this process for other variables as potential output variables does not yield complementary expressions, thereby invalidating the assumption for those cases.

\begin{algorithm}[t]
\caption{Pseudo Code of our Transformation Method}
\begin{algorithmic}[1]
    \STATE \textbf{Input:} A list of clauses $C$
    \STATE \textbf{Output:} List of primary outputs $PO$, primary inputs $PI$, intermediate variables $IV$, and Boolearn expressions $BE$
    \STATE $SC$ = [] \COMMENT{List of sub-clauses}
    \FOR{$i = 1$ to length($C$)}
        % \IF{$C[i] \cap SC = \emptyset$}
        %     \STATE Append \text{Simplify}(\text{FindBooleanExpression}([], $SC$)) to $BE$
        %     %\COMMENT{Simplify Boolean expression}
        %     \FOR{each item $w$ in $SC$}
        %         \IF{$w \notin IV$ and $w \neq v$}
        %             \STATE Append $w$ to $PI$
        %         \ENDIF
        %     \ENDFOR
        %     \STATE $SC$ = []
        % \ELSE
            \STATE Append $C[i]$ to $SC$
            \FOR{each item $v$ in $SC$}
                \IF{$v \notin PI$ and $v \notin IV$}
                    \STATE $f \gets \text{FindBooleanExpression}(v, SC)$ %\COMMENT{Find Boolean expression for $v$}
                    \STATE $g \gets \text{FindBooleanExpression}(\neg v, SC)$ %\COMMENT{Find Boolean expression for $\neg v$}
                    \IF{$f = \neg g$}
                        \STATE Append \text{Simplify}($f$) to $BE$ %\COMMENT{Simplify Boolean expression}
                        \IF{$f = True$ or $f = False$}
                            \STATE Append $v$ to $PO$
                        \ELSE
                            \STATE Append $v$ to $IV$
                        \ENDIF
                        \FOR{each item $w$ in $SC$}
                            \IF{$w \notin IV$ and $w \neq v$}
                                \STATE Append $w$ to $PI$
                            \ENDIF
                        \ENDFOR
                        \STATE SC = []
                        \STATE \textbf{break}
                    \ENDIF
                \ENDIF    
            \ENDFOR
        % \ENDIF
    \ENDFOR
    \STATE \textbf{Return} $PO, PI, IV, BE$
    \vspace{-0.65cm}
\end{algorithmic}

\label{alg1}
\end{algorithm}

During the transformation process, some sub-clauses may be under-specified, making them difficult to translate directly into a Boolean expression. A simple example is an OR function $x_3(x_1, x_2) = x_1 \vee x_2$, where the output is constrained to 1. The full description of the associated clauses would be $(\neg x_3 \vee x_1 \vee x_2) \wedge (x_3 \vee \neg x_1) \wedge (x_3 \vee \neg x_2) \wedge x_3$. However, since $x_3$ is equal to 1 in this case, these clauses can be simplified to $(x_1 \vee x_2)$. In this simplified form, where the output variable is not explicitly specified, we cannot directly extract the function with its output constrained to 1. To handle such cases, if the current clauses do not share variables with subsequent clauses, we simplify these clauses to a Boolean expression as the result of the transformation. The variables of these clauses are then treated as input variables to the simplified Boolean expression, and an auxiliary variable is assigned to the output, which is constrained to $1$.

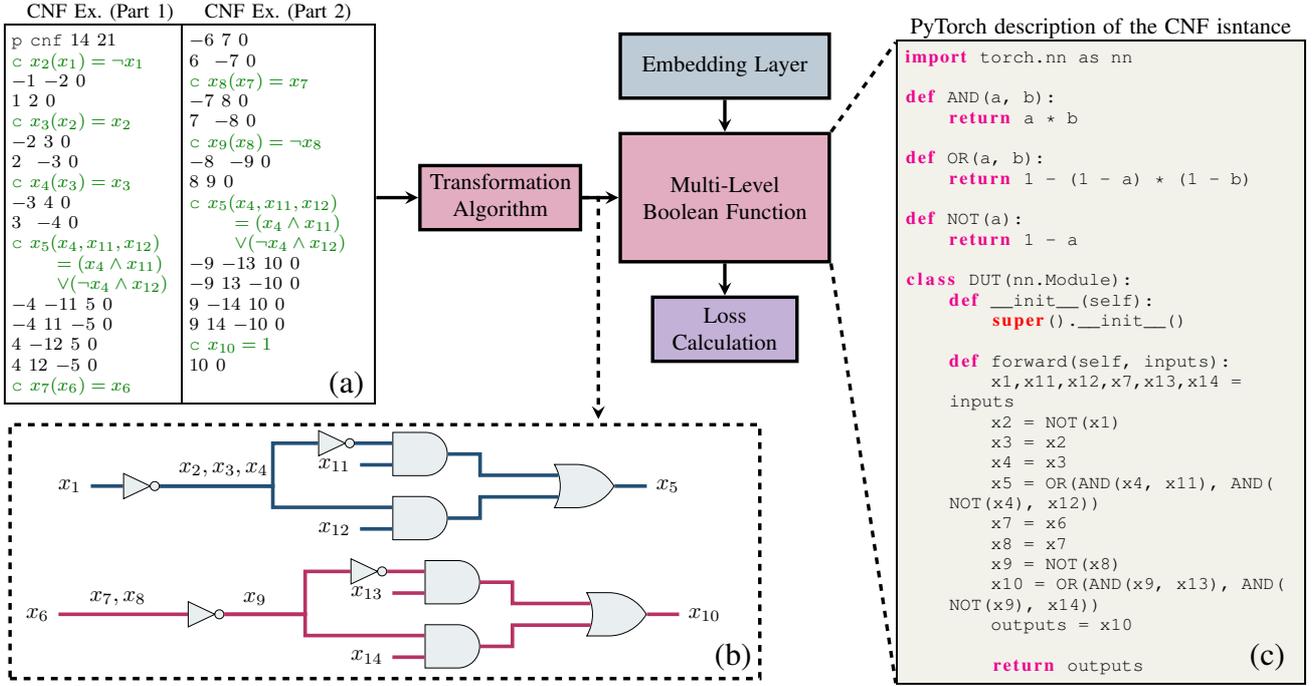
\begin{figure*}
    \centering
    \scalebox{0.85}{\begin{tikzpicture}[node distance=2cm,>=Latex]
    \node [ text width=2.75cm, align=center, line width=1.5pt] (code1) {
    \begin{minipage}[t]{\textwidth}
        \begin{lstlisting}[style=cnf, title=\small CNF Ex. (Part 1)]
$\text{p cnf}$ $14$ $21$
c $\textcolor{cnf-comment}{x_2(x_1) = \neg x_1}$
$-1$ $-2$ $0$ 
$1$ $2$ $0$ 
c $\textcolor{cnf-comment}{x_3(x_2) = x_2}$
$-2$ $3$ $0$ 
$2$  $-3$ $0$ 
c $\textcolor{cnf-comment}{x_4(x_3) = x_3}$
$-3$ $4$ $0$ 
$3$  $-4$ $0$ 
c $\textcolor{cnf-comment}{x_5(x_4,x_{11},x_{12})}$ $\textcolor{cnf-comment}{= (x_4 \wedge x_{11})}$ $\textcolor{cnf-comment}{\vee (\neg x_4 \wedge x_{12})}$
$-4$ $-11$ $5$ $0$ 
$-4$ $11$ $-5$ $0$ 
$4$ $-12$ $5$ $0$ 
$4$ $12$ $-5$ $0$ 
c $\textcolor{cnf-comment}{x_7(x_6) = x_6}$
\end{lstlisting}
\end{minipage}
    };
    \node[text width=2.75cm, align=center, line width=1.5pt, xshift=2.75cm] (code2) at (0,0) {
        \begin{minipage}[t]{\textwidth}
\begin{lstlisting}[style=cnf, title=\small CNF Ex. (Part 2)]
$-6$ $7$ $0$ 
$6$  $-7$ $0$ 
c $\textcolor{cnf-comment}{x_8(x_7) = x_7}$
$-7$ $8$ $0$ 
$7$  $-8$ $0$ 
c $\textcolor{cnf-comment}{x_9(x_8) = \neg x_8}$
$-8$  $-9$ $0$ 
$8$ $9$ $0$ 
c $\textcolor{cnf-comment}{x_5(x_4,x_{11},x_{12})}$ $\textcolor{cnf-comment}{= (x_4 \wedge x_{11})}$ $\textcolor{cnf-comment}{\vee (\neg x_4 \wedge x_{12})}$
$-9$ $-13$ $10$ $0$ 
$-9$ $13$ $-10$ $0$ 
$9$ $-14$ $10$ $0$ 
$9$ $14$ $-10$ $0$ 
c $\textcolor{cnf-comment}{x_{10} = 1}$
$10$ $0$
$~$
\end{lstlisting}
   \end{minipage}
    };

\node [draw,  fill = datemagenta!40, text width=2.25cm, minimum height=1cm, align=center, right=0.65cm of code2, line width=1.5pt] (transformer) {Transformation Algorithm};
\node [draw,  fill = datemagenta!40, text width=3cm, minimum height=2cm, align=center, right=3.75cm of code2, line width=1.5pt] (model) {Multi-Level Boolean Function};
\node [draw,  text width=3cm,fill = dateblue!30, minimum height=1cm, align=center, above=0.5cm of model, line width=1.5pt] (embedding) {Embedding Layer};
\node [draw, fill = Paired-9!40, text width=2cm, minimum height=1cm,align=center, below=0.5cm of model, line width=1.5pt] (loss) {Loss Calculation};

\node [text width=6.1cm, align=center, right=1.cm of model, yshift=-2.25cm] (parsed) { \lstset{style=mystyle} \begin{lstlisting}[language=Python, title=PyTorch description of the CNF isntance]
import torch.nn as nn

def AND(a, b):
    return a * b

def OR(a, b):
    return 1 - (1 - a) * (1 - b)

def NOT(a):
    return 1 - a

class DUT(nn.Module):
    def __init__(self):
        super().__init__()

    def forward(self, inputs):
        x1,x11,x12,x7,x13,x14 = inputs
        x2 = NOT(x1)
        x3 = x2
        x4 = x3
        x5 = OR(AND(x4, x11), AND(NOT(x4), x12))
        x7 = x6
        x8 = x7
        x9 = NOT(x8)
        x10 = OR(AND(x9, x13), AND(NOT(x9), x14))
        outputs = x10

        return outputs
\end{lstlisting}};

\draw [->, >=stealth, line width=1.5pt] (code2) -- node[above] {} (transformer);
    \draw [->, >=stealth, line width=1.5pt] (embedding.south) -- (model.north);
    \draw [->, >=stealth, line width=1.5pt] (transformer.east) -- (model.west);
    \draw [->, >=stealth, line width=1.5pt] (model.south) -- (loss.north);
    \draw [dashed, >=stealth, line width=1.5pt] (model.north east) -- ([shift={(0,-1cm)}]parsed.north west);
    \draw [dashed, >=stealth, line width=1.5pt] (model.south east) -- ([shift={(0,0.35cm)}]parsed.south west);

\node[not gate US, draw, logic gate inputs={n}, fill =  dateblue!10, scale = 1] (not_1) at (0.5,-4.5) {};
\node[and gate US, draw, logic gate inputs={nnn}, fill =  dateblue!10, scale = 1] (and_1) at ([xshift=4cm,yshift=0.5cm]not_1.output) {};
\node[and gate US, draw, logic gate inputs={nnn}, fill =  dateblue!10, scale = 1] (and_2) at ([xshift=4cm,yshift=-0.5cm]not_1.output) {};
\node[not gate US, draw, logic gate inputs={n}, fill =  dateblue!10, scale = 1] (not_2) at ([xshift=-1cm]and_1.input 1) {};
\node[or gate US, draw, logic gate inputs={nnn}, fill =  dateblue!10, scale = 1] (or_1) at ([xshift=6.5cm]not_1.output) {};

\node[not gate US, draw, logic gate inputs={n}, fill =  dateblue!10, scale = 1] (not_3) at (1.5,-6.5) {};
\node[and gate US, draw, logic gate inputs={nnn}, fill =  dateblue!10, scale = 1] (and_3) at ([xshift=3.5cm,yshift=0.5cm]not_3.output) {};
\node[and gate US, draw, logic gate inputs={nnn}, fill =  dateblue!10, scale = 1] (and_4) at ([xshift=3.5cm,yshift=-0.5cm]not_3.output) {};
\node[not gate US, draw, logic gate inputs={n}, fill =  dateblue!10, scale = 1] (not_4) at ([xshift=-1cm]and_3.input 1) {};
\node[or gate US, draw, logic gate inputs={nnn}, fill =  dateblue!10, scale = 1] (or_2) at ([xshift=6cm]not_3.output) {};

\draw [draw = dateblue, ultra thick] (not_1.input) -- ++(-0.5cm,0) node[left] {$x_1$};
\draw [draw = dateblue, ultra thick] (not_1.output)  node[shift={(0.15,0)}, above right] {$x_2, x_3, x_4$} -- ([xshift=1.75cm]not_1.output) |- (not_2.input);
\draw [draw = dateblue, ultra thick] (not_2.output) -- (and_1.input 1);
\draw [draw = dateblue, ultra thick] (not_1.output)  -- ([xshift=1.75cm]not_1.output) |- (and_2.input 1);
\draw [draw = dateblue, ultra thick] (and_1.output) -- ([xshift=0.5cm]and_1.output) |- (or_1.input 1);
\draw [draw = dateblue, ultra thick] (and_2.output) -- ([xshift=0.5cm]and_2.output) |- (or_1.input 3);
\draw [draw = dateblue, ultra thick] (and_1.input 3) -- ++(-0.5cm,0) node[left] {$x_{11}$};
\draw [draw = dateblue, ultra thick] (and_2.input 3) -- ++(-0.5cm,0) node[left] {$x_{12}$};
\draw [draw = dateblue, ultra thick] (or_1.output) -- ++(0.5cm,0) node[right] {$x_5$};

\draw  [draw = datemagenta, ultra thick] (not_3.input) node[shift={(-0.5,0)}, above left] {$x_7, x_8$} -- ++(-2cm,0) node[left] {$x_6$};
\draw   [draw = datemagenta, ultra thick](not_3.output) node[shift={(0.15,0)}, above right] {$x_9$} -- ([xshift=1.25cm]not_3.output) |- (not_4.input);
\draw   [draw = datemagenta, ultra thick](not_4.output) -- (and_3.input 1);
\draw   [draw = datemagenta, ultra thick](not_3.output)  -- ([xshift=1.25cm]not_3.output) |- (and_4.input 1);
\draw   [draw = datemagenta, ultra thick](and_3.output) -- ([xshift=0.5cm]and_3.output) |- (or_2.input 1);
\draw   [draw = datemagenta, ultra thick](and_4.output) -- ([xshift=0.5cm]and_4.output) |- (or_2.input 3);
\draw   [draw = datemagenta, ultra thick](and_3.input 3) -- ++(-0.5cm,0) node[left] {$x_{13}$};
\draw   [draw = datemagenta, ultra thick](and_4.input 3) -- ++(-0.5cm,0) node[left] {$x_{14}$};
\draw   [draw = datemagenta, ultra thick](or_2.output) -- ++(0.5cm,0) node[right] {$x_{10}$};

\draw [line width=1.5pt, dashed]([shift={(-1.25,0.25)}]not_1.north west) ++(-0.5,0.5) rectangle ([shift={(2.,-0.75)}]or_2.south east);
        \node[above left] at ([shift={(-4,3)}]or_2.north east) {\Large (a)};
        \node[above left] at ([shift={(2,-.75)}]or_2.south east) {\Large (b)};
        \node[above left] at ([shift={(10.25,-.75)}]or_2.south east) {\Large (c)};

\draw [-<, >=stealth, line width=1.5pt, dashed] (transformer.east) -- ([xshift=0.25cm]transformer.east) |- ([xshift=0.25cm,yshift=-3.25cm]transformer.east);

\end{tikzpicture}}
    \vspace{-0.25cm}
    \caption{The overall workflow of our sampling approach is illustrated including (a) a CNF example with comments highlighted in green, (b) its simplified multi-level, multi-output Boolean function in a circuit form for illustrative purposes with unconstrained and constrained paths highlighted in blue and red, respectively, and (c) its corresponding PyTorch description. }
    \label{fig1}
    \vspace{-0.5cm}
\end{figure*}

After transforming all the clauses, we integrate the resulting Boolean sub-expressions to construct a multi-level, multi-output Boolean function that is equisatisfiable with the original CNF. This function contains two types of paths: \textit{constrained paths} and \textit{unconstrained paths}. Constrained paths are those that run from primary inputs to primary outputs. The inputs along these paths must be adjusted to satisfy the explicit constraints applied to the primary output variables. Unconstrained paths originate from primary inputs and terminate at intermediate variables. Since they are not explicitly constrained to any values, any random initialization of their primary inputs will yield satisfying solutions for the variables on these paths. Fig. \ref{fig1}(a) and \ref{fig1}(b) present an example of a CNF and its corresponding circuit structure, produced through our transformation method. The figures highlight two types of paths: unconstrained paths in blue and constrained paths in red. In this example, any random assignment to the primary input variables $x_1$, $x_{11}$, and $x_{12}$ will satisfy the sub-clauses associated to the unconstrained path. Conversely, the primary input variables for the constrained path, namely $x_6$, $x_{13}$, and $x_{14}$, must be carefully selected such that $x_{10}$ is forced to be $1$.

With the definitions in place, the SAT solving/sampling problem now becomes a task of finding the primary inputs in the constrained paths to the multi-level, multi-output Boolean function resulting from the CNF transformation. Once the values of the primary input variables are determined to satisfy the constraints on the primary output variables, the intermediate variables can be computed using the corresponding Boolean operators in the function (see Fig. \ref{fig1}(c)).

To solve for the primary input variables in the constrained paths, we use GD optimization. Specifically, we reformulate the solving/sampling problem into a multi-output regression task, where the objective is to learn the primary input variables based on explicit constraints on the primary outputs and a differentiable model that maps the inputs to the outputs. While the multi-level, multi-output Boolean function fulfills the mapping requirements, it is not differentiable in its discrete form. To address this, we employ a probabilistic representation of logical operators \cite{Ardakani2017SC, NEURIPS2019_6562c5c1} such as AND, OR, NOT, XOR, and XNOR, enabling differentiability in the model as shown in Table \ref{tab1}. Such a representation allows us to relax the discrete operations into continuous ones.

\begin{table}
    \centering
    \caption{The probabilistic representation of logical operators.}
    \scalebox{0.77}{\begin{tabular}{c|c|c|c} 
\hline
\midrule
Operator & Input Variable & Output Variable & Derivative w.r.t Input\\ 
 \midrule
NOT & $P_{1}$ & $P_y = \overline{P_1} = 1 - P_{1}$ & $\dfrac{\partial P_y}{\partial P_1} = -1$  \\
\midrule
AND & $P_{1}$, $P_{2}$ & $P_y = P_{1}~P_{2}$ & $\dfrac{\partial P_y}{\partial P_1} = P_2$, $\dfrac{\partial P_y}{\partial P_2} = P_1$ \\
\midrule
OR & $P_{1}$, $P_{2}$ & $P_y = 1 - \overline{P_{1}}~\overline{P_{2}}$ & $\dfrac{\partial P_y}{\partial P_1} = \overline{P_{2}}$, $\dfrac{\partial P_y}{\partial P_2} = \overline{P_{1}}$\\
\midrule
% NAND & $P_{1}$, $P_{2}$ & $P_y = 1 - P_{1}~P_{2}$ & $\dfrac{\partial P_y}{\partial P_1} = -P_2$, $\dfrac{\partial P_y}{\partial P_2} = -P_1$\\
% \midrule
% NOR & $P_{1}$, $P_{2}$ & $P_y = \overline{P_{1}}~\overline{P_{2}}$ & $\dfrac{\partial P_y}{\partial P_1} = -\overline{P_{2}}$, $\dfrac{\partial P_y}{\partial P_2} = -\overline{P_{1}}$\\
% \midrule
XOR & $P_{1}$, $P_{2}$ & $P_y = \overline{P_{1}}~P_{2} + P_{1}~\overline{P_{2}}$ & $\dfrac{\partial P_y}{\partial P_1} = 1 -2P_2$, $\dfrac{\partial P_y}{\partial P_2} = 1 - 2 P_1$\\
\midrule
XNOR & $P_{1}$, $P_{2}$ & $P_y = P_{1}~P_{2} + \overline{P_{1}}~\overline{P_{2}}$ & $\dfrac{\partial P_y}{\partial P_1} = 2P_2-1$, $\dfrac{\partial P_y}{\partial P_2} = 2P_1-1$\\
\midrule
\hline

\end{tabular}}
    \label{tab1}
    \vspace{-0.5cm}
\end{table}

With the probabilistic model derived from replacing the discrete logical operators of the multi-level, multi-output Boolean function with their probabilistic counterparts, we can now learn satisfying primary input variables given the constraints on the output variables. To find satisfying solutions, we define the primary input variables a matrix $\textbf{V} \in \mathbb{R}^{b \times n}$, where $n$ indicates the number of primary input variables and $b$ denotes the batch size. Since our continuous multi-level, multi-output Boolean function takes real values between $0$ and $1$ in a form of probability, we convert the primary input variables into probabilities using the sigmoid function $\sigma(\cdot)$, such that:\vspace{-0.2cm}
\begin{equation}
    \textbf{P} = \sigma(\textbf{V}),\vspace{-0.1cm}
\end{equation}
where $\textbf{P} \in [0,1]^{b \times n}$ represents the input probabilities fed into the model. We refer to this conversion process as a form of continuous embedding, as it transforms the input space into a continuous probability space, allowing the model to interpret the inputs probabilistically. The model's primary outputs are then computed as:
\begin{equation}\vspace{-0.1cm}
    \textbf{Y} = \mathcal{F}(\textbf{P}),\vspace{-0.1cm}
\end{equation}
where $\mathcal{F}:[0,1]^{b \times n} \rightarrow [0,1]^{b \times m}$ represents the probabilistic multi-level, multi-output Boolean function. The matrix $\textbf{Y} \in [0,1]^{b \times m}$ holds the $m$ primary output variables across the $b$ data batches.

To optimize the inputs, we define an $\ell_2$-loss function $\mathcal{L}$ that measures the difference between the computed outputs $\textbf{Y}$ and the target output matrix $\textbf{T} \in {0,1}^{b \times m}$:\vspace{-0.2cm}
\begin{equation}
    \mathcal{L} = \sum_{b,m} \left|\left| \textbf{Y} - \textbf{T} \right|\right|^2_2.\vspace{-0.2cm}
\end{equation}
By minimizing this loss function through GD, the primary input variables (i.e., $\textbf{V}$) are iteratively updated. Upon convergence, the $b$ solutions are determined by converting the soft input values (i.e., $\textbf{V}$) into hard binary values (i.e., $\widetilde{\textbf{V}} \in {0, 1}^{b \times n}$). Among these $b$ solutions, only those that are satisfiable and non-redundant are retained as satisfiable solutions.

In gradient descent optimization, the objective is to compute the derivative of the loss function with respect to each primary input variable associated with the constrained paths. This can be achieved using the chain rule, leveraging the derivatives of the logical operators presented in Table \ref{tab1}. For instance, in the constrained path highlighted in red in Fig. \ref{fig1}(b), the derivative of the loss function with respect to the primary input variable $x_{13}$ can be expressed as follows:
\begin{equation}
    \dfrac{\partial \mathcal{L}}{\partial x_{13}} = \dfrac{\partial \mathcal{L}}{\partial x_{10}} \cdot \dfrac{\partial x_{10}}{\partial x_{13}} = 2\cdot (x_{10} - 1) \cdot (1 - x_9 \cdot x_{14})\cdot (1 - x_9).
\end{equation}
For simplicity, we will exclude the embedding process from our calculations in this example. With the computed derivative, the value of $x_{13}$ is updated using the following equation:
\begin{equation}
    x_{13} = x_{13} - \gamma \cdot \dfrac{\partial \mathcal{L}}{\partial x_{13}},
\end{equation}
where $\gamma$ represents the learning rate.

The overall workflow is illustrated in Fig. \ref{fig1}. Our method integrates a parser that describes the probabilistic multi-level, multi-output Boolean function  in PyTorch. Since each solution is processed and learned independent from others, our approach is highly parallelizable and benefits from GPU acceleration, enables fast and scalable sampling by processing multiple data batches simultaneously.

\begin{table*}[]
    \caption{The runtime performance of our sampling method against {\sc UniGen3}, {\sc CMSGen} and {\sc DiffSampler} in terms of unique solution throughput, where each sampler is tasked to generate a minimum of $1000$ solutions within a timeout (TO) of $2$ hours.}
    \vspace{-0.2cm}
    \centering
    \begin{tabular}{c|c|c|c|c|c|c|c|c} 
\hline
\midrule
SAT  & \# Primary & \# Primary  & \# Variables & \# Clauses & \multicolumn{4}{c}{Throughput (\# Unique Solutions per Second)}\\ 
Instance & Inputs & Outputs & (CNF) &  (CNF) & \textbf{This work} (Speedup) & {\sc UniGen3} & {\sc CMSGen} & {\sc DiffSampler}\\ 
% SAT  & \# Primary & \# Primary  & \# Variables & \# Clauses & \multirow{ 2}{*}{\textbf{This work} (Speedup)} & \multirow{ 2}{*}{{\sc UniGen3}} & \multirow{ 2}{*}{{\sc CMSGen}} & \multirow{ 2}{*}{{\sc DiffSampler}}\\ 
% Instance & Inputs & Outputs & (CNF) &  (CNF) & &  & &\\ 
 \midrule
or-$50$-$10$-$7$-UC-$10$ & $50$ & $4$ & $100$ & $254$ & $\textbf{5,974,780.8}~(79.6\times)$ & $64.7$ & $36,693.5$ & $75,040.1$ \\
% or-$50$-$10$-$7$-UC-$40$ & $50$ & $19$ & $100$ & $250$ & $\textbf{224,660.0}~(5.2\times)$ & $58.3$ & $43,555.0$ & $4,882.8$ \\
%  \midrule
or-$60$-$20$-$10$-UC-$10$ & $60$ & $5$ & $120$ & $305$ & $\textbf{4,777,137.7}~(86.0\times)$ & $81.7$ & $33,987.0$ & $55,521.3$ \\
% or-$60$-$20$-$10$-UC-$40$ & $60$ & $25$ & $120$ & $300$ & $\textbf{2,290,983.7}~(56.5\times)$ & $69.0$ & $40,572.1$ & $7,288.8$ \\
%  \midrule
or-$70$-$5$-$5$-UC-$10$ & $69$ & $7$ & $140$ & $357$ & $\textbf{2,468,613.4}~(77.8\times)$ & $94.5$ & $31,732.4$ & $16,035.1$ \\
% or-$70$-$5$-$5$-UC-$40$ & $69$ & $29$ & $140$ & $350$ & $\textbf{159,326.4}~(3.8\times)$ & $17.4$ & $41,731.9$ & $11,188.1$ \\
%  \midrule
or-$100$-$20$-$8$-UC-$10$ & $98$ & $10$ & $200$ & $510$ & $\textbf{1,707,142.3}~(51.6\times)$ & $43.4$ & $22,951.7$ & $33,175.3$ \\
% or-$100$-$20$-$8$-UC-$60$ & $98$ & $59$ & $200$ & $500$ & $\textbf{354,295.7}~(9\times)$ & $13.6$ & $26,401.4$ & $39,469.4$ \\
 \midrule
$75$-$10$-$1$-q & $83$ & $1$ & $452$ & $443$ & $\textbf{478,723.0}~(42.0\times)$ & $1.6$ & $11,281.8$ & $156.1$ \\
% $75$-$10$-$5$-q & $83$ & $1$ & $460$ & $720$ & $\textbf{1,790,007.7}~(175.1\times)$ & $1.1$ & $10,220.8$ & $373.4$ \\  
$75$-$10$-$10$-q & $79$ & $1$ & $456$ & $439$ & $\textbf{2,075,175.0}~(197.1\times)$ & $1.6$ & $10,527.4$ & $251.8$ \\ 
% \midrule
$90$-$10$-$1$-q & $51$ & $1$ & $432$ & $411$ & $\textbf{2,809,981.5}~(251.7\times)$ & $1.0$ & $11,162.5$ & $227.9$ \\
% $90$-$10$-$5$-q & $45$ & $1$ & $460$ & $720$ & $\textbf{2,644,714.0}~(224.7\times)$ & $2.2$ & $11,771.8$ & $58.5$  \\
$90$-$10$-$10$-q & $31$ & $1$ & $428$ & $391$ & $\textbf{3,567,035.2}~(326.9\times)$ & $1.4$ & $10,913.0$ & $57.9$\\
\midrule
s$15850$a\_$3$\_$2$ & $600$ & $3$ & $10,908$ & $24,476$ & $\textbf{20,267.1}~(47.1\times)$ & $0.4$ & $430.4$ & TO \\
s$15850$a\_$7$\_$4$ & $600$ & $7$ & $10,926$ & $24,552$ & $\textbf{14,930.5}~(34.1\times)$ & $0.5$ & $437.9$ &  TO \\
s$15850$a\_$15$\_$7$ & $600$ & $15$ &  $10,995$ & $24,836$ & $\textbf{14,177.1}~(33.6\times)$ & $0.5$ & $422.2$ & TO \\
\midrule
Prod-$8$  & $293$ & $2$ & $14,952$ & $74,702$ & $\textbf{994.9}~(523.6\times)$ & $1.9$ & $0.2$ & TO \\
Prod-$20$ & $677$ & $2$ & $37,320$ & $186,734$ & $\textbf{139.1}~(347.8\times)$ & $0.4$ & TO &  TO \\
Prod-$32$ & $1061$ & $2$ & $59,688$ & $298,766$ & $\textbf{96.0}~(480\times)$ & $0.2$ & TO &  TO \\
\midrule
\hline
\end{tabular}

    \label{tab2}
    \vspace{-0.5cm}
\end{table*}

\vspace{-0.2cm}
\section{Experimental Results}
In this section, we demonstrate the effectiveness of our sampling technique. To accomplish this, we developed a prototype of our approach using PyTorch, an open-source machine learning library that integrates Torch's powerful GPU-optimized backend with a Python-friendly interface. For a thorough assessment, we evaluated $60$ problem instances of varying sizes from a public domain benchmark suite. The experiments were performed on a system featuring an Intel Xeon E5-2698 processor running at $2.2$ GHz and $8$ NVIDIA V$100$ GPUs, each with $32$ GB of memory. We present the runtime performance of our method in terms of throughput, defined as the number of unique and valid solutions generated per second, using a single NVIDIA V1$00$ GPU. GD was employed as the optimizer during the training phase, with the learning rate set to $10$ and the number of iterations to $5$. We varied the batch size between $100$ and $1,000,000$, based on the specific instances tested.

We compare the performance of our sampling method with leading SAT sampling methods, specifically {\sc UniGen3} \cite{Soos2020unigen3}, {\sc CMSGen} \cite{Golia2021cmsgen}, and {\sc DiffSampler} \cite{Ardakani2024diffsampler}. These samplers operate directly on the CNF of SAT instances, whereas our method handles the simplified multi-level, multi-output Boolean expressions derived from transforming their logical constraints. Both {\sc UniGen3} and {\sc CMSGen} are highly optimized implementations written in C++, while {\sc DiffSampler} is a Python-based, GPU-accelerated SAT sampler built using the high-performance JAX library. {\sc UniGen3} and {\sc CMSGen} were tested on a server-grade Intel Xeon Gold $6134$ CPU with $3.2$ GHz clock rate and 1TB of RAM. {\sc DiffSampler} was run on a system featuring an Intel Xeon E$5$-$2698$ processor at $2.2$ GHz and $8$ NVIDIA V$100$ GPUs, each equipped with $32$ GB of memory.

\begin{figure}[t]
    \centering
    \input{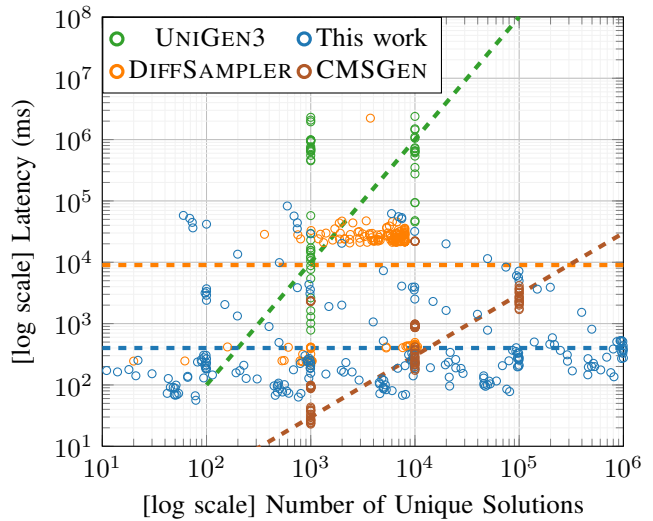}
    \caption{A log-log plot showing the runtime in milliseconds versus the number of unique satisfying solutions across all $60$ instances from the sampling benchmark. The dotted lines represent the performance trends for each sampler.}
    \label{fig2}
    \vspace{-0.5cm}
\end{figure}

\vspace{-0.2cm}
\subsection{Runtime Performance}
Table \ref{tab2} presents the sampling performance of our method in terms of throughput for a representative subset of $14$ instances from the sampling benchmark. Throughput is defined as the number of unique solutions generated per second. Each sampler is required to produce at least $1,000$ solutions within a maximum runtime of $2$ hours. The table shows the best throughput results obtained from each sampler. Our method consistently outperforms state-of-the-art samplers in unique solution throughput, achieving speedups ranging from $33.6\times$ to $523.6\times$, depending on the under-test SAT instances. This substantial performance improvement is due to two key factors. First, many logical constraints in the CNF representation are satisfied during the transformation, where sub-clauses are converted into simplified Boolean sub-expressions. This transforms the SAT sampling task into solving constrained paths in a simplified, multi-level, multi-output Boolean expression, significantly reducing the number of logical operations. Second, by framing the sampling problem as a learning task, where the computation of each sample is independent, GPU acceleration can be leveraged to further enhance runtime performance.

Fig. \ref{fig2} demonstrates how the runtime performance of each sampler varies as the number of unique solutions increases. A critical aspect of this comparison is the overall efficiency of our sampling method relative to {\sc UniGen3}, {\sc CMSGen}, and {\sc DiffSampler}. This is particularly evident when sampling larger quantities of solutions, where the runtime shows only a slight increase as the solution count grows.

\vspace{-0.2cm}
\subsection{Learning Dynamics}
We analyze the learning dynamics of our sampling method, focusing on hyper-parameters such as iterations and batch size. Fig. \ref{fig3} shows the progress in generating unique solutions over $10$ iterations, where the number of unique solutions increases with more iterations. Increasing the batch size improves runtime performance by leveraging GPU parallelism, but at the cost of higher memory usage. The GPU memory demand, as shown in Fig. \ref{fig3}, grows with both the complexity of the Boolean function derived from the CNF transformation and the batch size. In scenarios requiring a high number of unique samples with limited GPU memory, the practical solutions are to either run more iterations, reducing throughput, or use GPUs with larger memory.

% We present an in-depth analysis of the learning dynamics of our sampling method, focusing on hyper-parameters like the number of iterations and batch size. Fig. \ref{fig3} illustrates our sampler's progress in generating unique solutions over the course of $10$ training iterations. The learning curves indicate that as the number of iterations increases, our sampler uncovers more unique solutions. Although gradient descent lacks a theoretical guarantee of reaching the global minimum in non-convex landscapes, including our continuous SAT problem formulation, our experiments show its ability to find high-performing solutions in this context. While these solutions may not represent the global minimum in the continuous domain, they still meet the logical constraints of the discrete form.

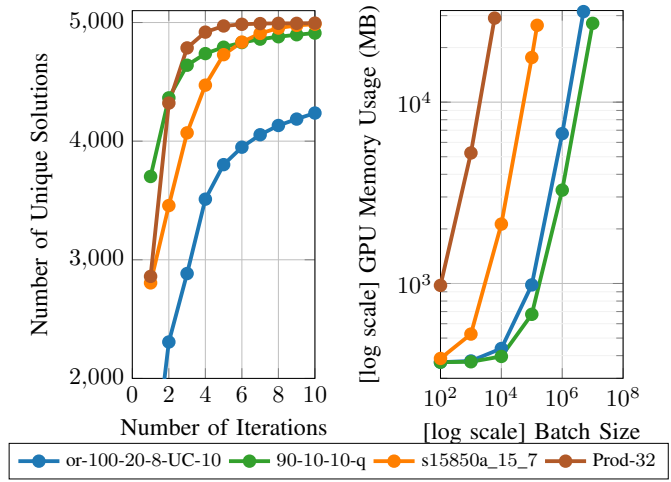
\begin{figure}
    \centering
    \scalebox{0.9}{
    \begin{tikzpicture}
\hspace{-4.5cm}
    \begin{axis}[
        xlabel={Number of Iterations},
        ylabel={Number of Unique Solutions},
        height=7cm, width=4.25cm,
        xmin=0, xmax=10,
        ymin=2000, ymax=5100,
        % ymode=log,
        grid=both,
        grid style={line width=.1pt, draw=gray!10},
        major grid style={line width=.2pt,draw=gray!50},
        xlabel near ticks,
        ylabel near ticks,
        tick align=inside,
        legend columns=1,
        % transpose legend,
        legend style={at={(1,0)},anchor=south east},
    ]
    % \addplot[ultra thick, mark=*, color=Paired-1] coordinates {
    %     (1, 191)
    %     (2, 377)
    %     (3, 610)
    %     (4, 877)
    %     (5, 1120)
    %     (6, 1354)
    %     (7, 1585)
    %     (8, 1793)
    %     (9, 1989)
    %     (10, 2202)
    % };
    % \addlegendentry{$75$-$10$-$1$-q}

    % \addplot[ultra thick, mark=*, color=Paired-2] coordinates {
    %     (1, 3974)
    %     (2, 4276)
    %     (3, 4452)
    %     (4, 4558)
    %     (5, 4666)
    %     (6, 4713)
    %     (7, 4773)
    %     (8, 4825)
    %     (9, 4853)
    %     (10, 4868)
    % };
    % \addlegendentry{$75$-$10$-$5$-q}
    
    % \addplot[ultra thick, mark=*, color=Paired-1] coordinates {
    %     (1, 1426)
    %     (2, 2306)
    %     (3, 3071)
    %     (4, 3632)
    %     (5, 4000)
    %     (6, 4281)
    %     (7, 4436)
    %     (8, 4537)
    %     (9, 4634)
    %     (10, 4690)
    % };
    % \addlegendentry{$75$-$10$-$10$-q}

    \addplot[ultra thick, mark=*, color=Paired-1] coordinates {
        (1, 1015)
        (2, 2307)
        (3, 2884)
        (4, 3511)
        (5, 3802)
        (6, 3949)
        (7, 4054)
        (8, 4132)
        (9, 4186)
        (10, 4237)
    };
    % \addlegendentry{or-$100$-$20$-$8$-UC-$10$}
    
    % \addplot[ultra thick, mark=*, color=Paired-4] coordinates {
    %     (1, 4232)
    %     (2, 4492)
    %     (3, 4653)
    %     (4, 4738)
    %     (5, 4798)
    %     (6, 4842)
    %     (7, 4871)
    %     (8, 4891)
    %     (9, 4907)
    %     (10, 4918)
    % };
    % \addlegendentry{$90$-$10$-$1$-q}
    
    % \addplot[ultra thick, mark=*, color=Paired-5] coordinates {
    %     (1, 4508)
    %     (2, 4813)
    %     (3, 4922)
    %     (4, 4953)
    %     (5, 4970)
    %     (6, 4984)
    %     (7, 4989)
    %     (8, 4990)
    %     (9, 4990)
    %     (10, 4993)
    % };
    % \addlegendentry{$90$-$10$-$5$-q}

    \addplot[ultra thick, mark=*, color=Paired-3] coordinates {
        (1, 3702)
        (2, 4365)
        (3, 4640)
        (4, 4738)
        (5, 4792)
        (6, 4831)
        (7, 4860)
        (8, 4880)
        (9, 4897)
        (10, 4911)
    };
    % \addlegendentry{$90$-$10$-$10$-q}

    % \addplot[ultra thick, mark=*, color=Paired-7] coordinates {
    %     (1, 3610)
    %     (2, 4560)
    %     (3, 4868)
    %     (4, 4971)
    %     (5, 4987)
    %     (6, 4997)
    %     (7, 4997)
    %     (8, 4999)
    %     (9, 4999)
    %     (10, 5000)
    % };
    % \addlegendentry{s$15850$a\_$3$\_$2$}

    % \addplot[ultra thick, mark=*, color=Paired-8] coordinates {
    %     (1, 3426)
    %     (2, 4189)
    %     (3, 4615)
    %     (4, 4831)
    %     (5, 4924)
    %     (6, 4968)
    %     (7, 4988)
    %     (8, 4993)
    %     (9, 4997)
    %     (10, 4997)
    % };
    % \addlegendentry{s$15850$a\_$7$\_$4$}

    \addplot[ultra thick, mark=*, color=Paired-7] coordinates {
        (1, 2805)
        (2, 3457)
        (3, 4070)
        (4, 4472)
        (5, 4729)
        (6, 4835)
        (7, 4908)
        (8, 4953)
        (9, 4976)
        (10, 4986)
    };
    % \addlegendentry{s$15850$a\_$15$\_$7$}

    % \addplot[ultra thick, mark=*, color=Paired-10] coordinates {
    %     (1, 2724)
    %     (2, 4260)
    %     (3, 4754)
    %     (4, 4905)
    %     (5, 4961)
    %     (6, 4983)
    %     (7, 4989)
    %     (8, 4991)
    %     (9, 4992)
    %     (10, 4992)
    % };
    % \addlegendentry{Prod-$8$}
    % \addplot[ultra thick, mark=*, color=Paired-10] coordinates {
    %     (1, 2764)
    %     (2, 4288)
    %     (3, 4757)
    %     (4, 4928)
    %     (5, 4973)
    %     (6, 4985)
    %     (7, 4991)
    %     (8, 4993)
    %     (9, 4996)
    %     (10, 4998)
    % };
    % \addlegendentry{Prod-$20$}
    \addplot[ultra thick, mark=*, color=Paired-11] coordinates {
        (1, 2860)
        (2, 4321)
        (3, 4786)
        (4, 4919)
        (5, 4971)
        (6, 4985)
        (7, 4991)
        (8, 4993)
        (9, 4993)
        (10, 4995)
    };
    % \addlegendentry{Prod-$32$}

    \end{axis}
\hspace{4.5cm}
\begin{axis}[
        xlabel={[log scale] Batch Size},
        ylabel={[log scale] GPU Memory Usage (MB)},
        height=7cm, width=4.25cm,
        xmin=100, xmax=100000000,
        ymin=300, ymax=32000,
        xmode=log,
        ymode=log,
        grid=both,
        grid style={line width=.1pt, draw=gray!10},
        major grid style={line width=.2pt,draw=gray!50},
        xlabel near ticks,
        ylabel near ticks,
        tick align=inside,
        legend columns=4,
        legend style={font=\footnotesize, at={(1.25,-0.275)},anchor=south east},
    ]
    % \addplot[ultra thick, mark=*, color=Paired-1] coordinates {
    %     (100, 368)
    %     (1000, 372)
    %     (10000, 414)
    %     (100000, 844)
    %     (500000, 4944)
    % };
    % \addlegendentry{$75$-$10$-$1$-q}

    % \addplot[ultra thick, mark=*, color=Paired-2] coordinates {
    %     (100, 368)
    %     (1000, 372)
    %     (10000, 414)
    %     (100000, 842)
    %     (1000000, 4980)
    % };
    % \addlegendentry{$75$-$10$-$5$-q}
    
    % \addplot[ultra thick, mark=*, color=Paired-1] coordinates {
    %     (100, 368)
    %     (1000, 372)
    %     (10000, 414)
    %     (100000, 860)
    %     (1000000, 5052)
    % };
    % \addlegendentry{$75$-$10$-$10$-q}

    \addplot[ultra thick, mark=*, color=Paired-1] coordinates {
        (100, 368)
        (1000, 374)
        (10000, 438)
        (100000, 982)
        (1000000, 6706)
        (5000000, 31580)
    };
    \addlegendentry{or-$100$-$20$-$8$-UC-$10$}
    
    % \addplot[ultra thick, mark=*, color=Paired-4] coordinates {
    %     (100, 368)
    %     (1000, 370)
    %     (10000, 402)
    %     (100000, 710)
    %     (1000000, 3732)
    % };
    % \addlegendentry{$90$-$10$-$1$-q}
    
    % \addplot[ultra thick, mark=*, color=Paired-5] coordinates {
    %     (100, 368)
    %     (1000, 370)
    %     (10000, 402)
    %     (100000, 726)
    %     (500000, 3832)
    % };
    % \addlegendentry{$90$-$10$-$5$-q}

    \addplot[ultra thick, mark=*, color=Paired-3] coordinates {
        (100, 368)
        (1000, 370)
        (10000, 396)
        (100000, 676)
        (1000000, 3272)
        (10000000, 27208)
    };
    \addlegendentry{$90$-$10$-$10$-q}

    % \addplot[ultra thick, mark=*, color=Paired-10] coordinates {
    %     (100, 388)
    %     (1000, 530)
    %     (10000, 2128)
    %     (100000, 17878)
    %     (150000, 27040)
    % };
    % \addlegendentry{s$15850$a\_$3$\_$2$}

    % \addplot[ultra thick, mark=*, color=Paired-10] coordinates {
    %     (100, 386)
    %     (1000, 524)
    %     (10000, 2094)
    %     (100000, 17284)
    %     (150000, 26064)
    % };
    % \addlegendentry{s$15850$a\_$7$\_$3$}
    \addplot[ultra thick, mark=*, color=Paired-7] coordinates {
        (100, 386)
        (1000, 526)
        (10000, 2126)
        (100000, 17632)
        (150000, 26532)
    };
    \addlegendentry{s$15850$a\_$15$\_$7$}

% \addplot[ultra thick, mark=*, color=Paired-7] coordinates {
    %     (100, 520)
    %     (1000, 1588)
    %     (10000, 12980)
    %     (20000, 25497)
    % };
    % \addlegendentry{Prod-$8$}

    % \addplot[ultra thick, mark=*, color=Paired-8] coordinates {
    %     (100, 748)
    %     (1000, 3416)
    %     (10000, 31888)
    % };
    % \addlegendentry{Prod-$20$}

    \addplot[ultra thick, mark=*, color=Paired-11] coordinates {
        (100, 976)
        (1000, 5246)
        (6000, 29082)
    };
    \addlegendentry{Prod-$32$}

    \end{axis}
    
\end{tikzpicture}}
    \caption{(Left) Learning curve showing the number of unique satisfying solutions over iterations. (Right) Log-log plot of GPU memory usage (MB), measured by ``nvidia-smi'', across different batch sizes for a subset of $4$ CNF instances.}
    \label{fig3}
    \vspace{-0.4cm}
\end{figure}

% \begin{figure}
%     \centering
%     \input{fig4}
%     \caption{Log-log plot of GPU memory usage of our sampler in megabytes, measured by ``nvidia-smi'' across different batch sizes for a representative subset of $4$ instances from the sampling benchmark.}
%     \label{fig4}
% \end{figure}

% One way to enhance the runtime performance of our sampler is by increasing the batch size, enabling more samples to be generated simultaneously through the parallel processing capabilities of GPUs. However, this improvement comes at the expense of GPU memory usage. The GPU memory demand rises based on the complexity of the simplified multi-output, multi-level Boolean function derived from the CNF transformation and the batch size. For instance, Fig. \ref{fig3} displays GPU memory usage, as recorded by ``nvidia-smi'', across varying batch sizes. The figure highlights the significant increase in memory consumption with larger batch sizes. In cases where the goal is to generate a high number of unique samples but GPU memory is limited, the practical solution is to either run more iterations, resulting in lower throughput, or use more powerful GPUs with greater memory capacity.
\vspace{-0.2cm}
\subsection{Ablation Study}
In this section, we analyze the extent of GPU acceleration by comparing the runtime performance of our sampler between the GPU and CPU, as presented in Fig. \ref{fig5}. The data shows that GPU execution results in an average speedup of $6.8\times$ over CPU execution. Additionally, we provide the rate of reduction in the number of bit-wise operations due to the transformation, measured as the number of operations in the CNF divided by the number of operations in the resulting multi-level, multi-output Boolean function in terms of 2-input gate equivalents in Fig. \ref{fig5}, demonstrating an average reduction of $4.2\times$. Finally, we present the transformation time required to convert CNF into a multi-level, multi-output Boolean function in Fig. \ref{fig5}. This conversion time is comparable to that of conversion time of SAT applications represented in higher-level logical formats into CNF. Given the superior runtime performance of our sampler, we suggest that SAT applications in high-level logical formats could be directly transformed into a multi-level, multi-output Boolean function.

\vspace{-0.2cm}
\section{Related Work}

Numerous SAT formula sampling methods have been explored in prior research. For example, {\sc UniGen3} provides approximate uniformity guarantees \cite{yash2022barbarik}, while both {\sc CMSGen} and {\sc Quicksampler} \cite{dutra2018quicksampler} emphasize sampling efficiency. Other studies have also examined the use of data-parallel hardware for SAT solving, primarily focusing on parallelizing CDCL and various heuristic-based algorithms \cite{costa2013parallelization, osama2021sat}. Some recent efforts, such as {\sc MatSat} \cite{sato2021matsat} and {\sc NeuroSAT} \cite{amizadeh2018learning}, have attempted to frame SAT instances as constrained numerical optimization problems. However, these methods have struggled to demonstrate the effectiveness of GPU-accelerated formula sampling on large and diverse standard benchmarks, typically focusing on small, random instances. Recently, a new differentiable sampling technique called {\sc DiffSampler} was proposed in \cite{Ardakani2024diffsampler}, enabling GPU-accelerated SAT sampling on standard benchmarks and achieving competitive runtime performance compared to {\sc UniGen3} and {\sc CMSGen}. {\sc Demotic} is another GPU-accelerated, differentiable sampler specifically designed for CircuitSAT problems in CRV. It operates directly on circuit structures described in hardware description languages such as Verilog \cite{Ardakani2025Demotic}.

There have been only a few attempts in the literature that focus solely on extracting circuit structure from CNF descriptions \cite{Fu2007ExtractingLC, roy2004restoring}, primarily to recover information lost during the Tseitin transformation. These approaches typically rely on pattern matching based on predefined gate structures. In contrast, our transformation method is more general, capable of restoring any combination of logical gates from CNF, and, more importantly, it enables high-throughput SAT sampling using GPUs.

\begin{figure}[t]
    \centering
    \scalebox{0.85}{
    \begin{tikzpicture}
\hspace{-7cm}
  \begin{axis}[
        ybar, axis on top,
        height=3cm, width=8cm,
        bar width=0.2cm,
        ymajorgrids, tick align=inside,
        major grid style={draw =white},
        enlarge y limits={value=.1,upper},
        ymin=0, ymax=15,
        axis x line*=bottom,
        axis y line*=left,
        y axis line style={opacity=0},
        tickwidth=0pt,
        enlarge x limits={abs = 1.cm},
        x = 0.01cm,
        legend style={
            at={(0.5,-0.3)},
            anchor=north,
            legend columns=3,
            cells={align=left},
            /tikz/every even column/.append style={column sep=0.3cm}
        },
        every node near coord/.append style={font=\footnotesize},
        ylabel={GPU Speedup},
        ylabel style={yshift=-15pt},
        symbolic x coords={
         CNF Instances  },
       xtick=data,
       nodes near coords=\rotatebox{90}{
        \pgfmathprintnumber[precision=1]{\pgfplotspointmeta}$\times$
       }
    ]
   %  \addplot [draw=none, fill=Paired-1!60] coordinates {
      
   %    (CNF Instances, 11.1)};
   % \addplot [draw=none,fill=Paired-3!60] coordinates {
   %    (CNF Instances, 11.0)};
   \addplot [draw=none,fill=Paired-1] coordinates {
      (CNF Instances,11.9)};
   %  \addplot [draw=none,fill=Paired-7!60] coordinates {
   %    (CNF Instances,8.9)};
   % \addplot [draw=none,fill=Paired-9!60] coordinates {
   %    (CNF Instances,8.5)};   
    \addplot [draw=none,fill=Paired-3] coordinates {
      (CNF Instances,8.1)};
    % \addplot [draw=none,fill=Paired-11!60] coordinates {
    %   (CNF Instances,4.0)};
    % \addplot [draw=none,fill=Paired-11!60] coordinates {
    %   (CNF Instances,3.7)};
    \addplot [draw=none,fill=Paired-7] coordinates {
      (CNF Instances,4.5)};
    % \addplot [draw=none,fill=Paired-11!60] coordinates {
    %   (CNF Instances,2.7)};
    % \addplot [draw=none,fill=Paired-11!60] coordinates {
    %   (CNF Instances,1.9)};
    \addplot [draw=none,fill=Paired-11] coordinates {
      (CNF Instances,2.5)};

    % \legend{$75$-$10$-$10$-q, $90$-$10$-$10$-q, s$15850$a\_$15$\_$7$,  Prod-$32$}
     % \legend{$75$-$10$-$1$-q, $75$-$10$-$5$-q, $75$-$10$-$10$-q, $90$-$10$-$1$-q, $90$-$10$-$5$-q, $90$-$10$-$10$-q, s$15850$a\_$3$\_$2$, s$15850$a\_$7$\_$5$, s$15850$a\_$15$\_$7$, Prod-$8$, Prod-$20$, Prod-$32$}
  \end{axis}

\hspace{3.5cm}
\begin{axis}[
        ybar, axis on top,
        height=3cm, width=8cm,
        bar width=0.2cm,
        ymajorgrids, tick align=inside,
        major grid style={draw =white},
        enlarge y limits={value=.1,upper},
        ymin=0, ymax=5,
        axis x line*=bottom,
        axis y line*=left,
        y axis line style={opacity=0},
        tickwidth=0pt,
        enlarge x limits={abs = 1.cm},
        x = 0.01cm,
        legend style={
            at={(0.5,-0.3)},
            anchor=north,
            legend columns=3,
            cells={align=left},
            /tikz/every even column/.append style={column sep=0.3cm}
        },
        every node near coord/.append style={font=\footnotesize},
        ylabel={Ops Reduction},
        ylabel style={yshift=-15pt},
        symbolic x coords={
         CNF Instances  },
       xtick=data,
       nodes near coords=\rotatebox{90}{
        \pgfmathprintnumber[precision=1]{\pgfplotspointmeta}$\times$
       }
    ]
   %  \addplot [draw=none, fill=Paired-1!60] coordinates {
      
   %    (CNF Instances, 11.1)};
   % \addplot [draw=none,fill=Paired-3!60] coordinates {
   %    (CNF Instances, 11.0)};
   \addplot [draw=none,fill=Paired-1] coordinates {
      (CNF Instances,4.3)};
   %  \addplot [draw=none,fill=Paired-7!60] coordinates {
   %    (CNF Instances,8.9)};
   % \addplot [draw=none,fill=Paired-9!60] coordinates {
   %    (CNF Instances,8.5)};   
    \addplot [draw=none,fill=Paired-3] coordinates {
      (CNF Instances,4.2)};
    % \addplot [draw=none,fill=Paired-11!60] coordinates {
    %   (CNF Instances,4.0)};
    % \addplot [draw=none,fill=Paired-11!60] coordinates {
    %   (CNF Instances,3.7)};
    \addplot [draw=none,fill=Paired-7] coordinates {
      (CNF Instances,4.5)};
    % \addplot [draw=none,fill=Paired-11!60] coordinates {
    %   (CNF Instances,2.7)};
    % \addplot [draw=none,fill=Paired-11!60] coordinates {
    %   (CNF Instances,1.9)};
    \addplot [draw=none,fill=Paired-11] coordinates {
      (CNF Instances,3.6)};

    % \legend{$75$-$10$-$10$-q, $90$-$10$-$10$-q, s$15850$a\_$15$\_$7$,  Prod-$32$}
     % \legend{$75$-$10$-$1$-q, $75$-$10$-$5$-q, $75$-$10$-$10$-q, $90$-$10$-$1$-q, $90$-$10$-$5$-q, $90$-$10$-$10$-q, s$15850$a\_$3$\_$2$, s$15850$a\_$7$\_$5$, s$15850$a\_$15$\_$7$, Prod-$8$, Prod-$20$, Prod-$32$}
  \end{axis}

  \hspace{3.5cm}
  \begin{axis}[
        ybar, axis on top,
        height=3cm, width=8cm,
        bar width=0.2cm,
        ymajorgrids, tick align=inside,
        major grid style={draw =white},
        enlarge y limits={value=.1,upper},
        ymin=0, ymax=300,
        axis x line*=bottom,
        axis y line*=left,
        y axis line style={opacity=0},
        tickwidth=0pt,
        enlarge x limits={abs = 1.cm},
        x = 0.01cm,
        legend style={
            at={(-1.5,-0.4)},
            anchor=north,
            legend columns=4,
            cells={align=left},
            /tikz/every even column/.append style={column sep=0.3cm}
        },
        every node near coord/.append style={font=\footnotesize},
        ylabel={Latency (s)},
        ylabel style={yshift=-10pt},
        symbolic x coords={
         CNF Instances  },
       xtick=data,
       nodes near coords=\rotatebox{90}{
        \pgfmathprintnumber[precision=1]{\pgfplotspointmeta}
       }
    ]
   \addplot [draw=none,fill=Paired-1] coordinates {
      (CNF Instances,2.1)};
    \addplot [draw=none,fill=Paired-3] coordinates {
      (CNF Instances,22.8)};
    \addplot [draw=none,fill=Paired-7] coordinates {
      (CNF Instances,31.6)};
    \addplot [draw=none,fill=Paired-11] coordinates {
      (CNF Instances,292.2)};

    \legend{or-$100$-$20$-$8$-UC-$10$, $90$-$10$-$10$-q, s$15850$a\_$15$\_$7$,  Prod-$32$}
  \end{axis}

  \end{tikzpicture}}
    \caption{Comparison of GPU speedup over CPU (left), reduction rate of bit-wise operations in $2$-input gate equivalents (middle), and transformation time from CNF to a simplified multi-level, multi-output Boolean function (right) for a subset of $4$ instances.}
    \label{fig5}
    \vspace{-0.3cm}
\end{figure}
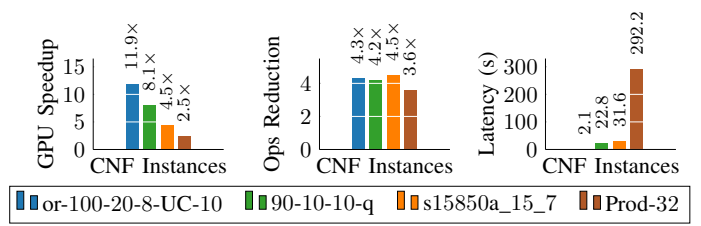
\section{Conclusion}
In this paper, we have introduced a novel GPU-accelerated approach for SAT sampling that significantly outperforms traditional methods. By transforming CNF representations into simplified multi-level, multi-output Boolean functions and employing gradient-based optimization, our method reinterprets SAT as a supervised multi-output regression task. This enables parallel, bit-wise operations across tensor elements, leading to substantial runtime improvements. With speedups ranging from $33.6\times$ to $523.6\times$ compared to existing heuristic samplers, our extensive evaluation on $60$ benchmark instances demonstrates the effectiveness and efficiency of this new technique for SAT sampling. This performance gain is primarily attributed to the GPU's acceleration over CPU execution and the reduction in the number of logic operations resulting from our transformation.
% \bibliography{references, references_scheduling, references_accel, references_algo}

\bibliographystyle{IEEEtran}
\bibliography{sat_sampling.bib}

\end{document}